\documentclass[runningheads]{llncs}

\usepackage[T1]{fontenc}
\usepackage{graphicx}
\usepackage{amsmath}
\usepackage{amssymb}
\usepackage{booktabs}
\usepackage{tabularx}
\usepackage{url}

\usepackage{docmute}

\let\realbibliographystyle\bibliographystyle
\let\realbibliography\bibliography
\renewcommand{\bibliographystyle}[1]{}
\renewcommand{\bibliography}[1]{}

\begin{document}

% ==================================================
% Main paper
% ==================================================

%
\title{An Attention Mechanism for Robust Multimodal Integration in a Global Workspace Architecture\thanks{Accepted at ICANN 2026.}}
\titlerunning{An Attention Mechanism for Robust Multimodal Integration }
% If the paper title is too long for the running head, you can set
% an abbreviated paper title here
%
\author{
Roland Bertin-Johannet\inst{1}\orcidID{0009-0007-4883-2999}\thanks{Corresponding author: \email{[roland.bertin-johannet@proton.me]}}
\and Lara Scipio\inst{1}
\and Leopold Maytié\inst{1}
\and Rufin VanRullen\inst{1}
}

\authorrunning{R. Bertin-Johannet et al.}

\institute{
CerCo, CNRS, Université de Toulouse; ANITI, Artificial and Natural Intelligence Toulouse Institute, Toulouse, France
}

%
% First names are abbreviated in the running head.
% If there are more than two authors, 'et al.' is used.
%
\maketitle              % typeset the header of the contribution
\begin{abstract}

%Robust multimodal systems must remain effective when some modalities are noisy, degraded, or unreliable. Global Workspace Theory (GWT), inspired by cognitive neuroscience, suggests that such flexibility could arise via attentional selection of the most relevant modalities. While recent implementations of GWT have explored multimodal representation capabilities, the role of top-down modality selection within such systems remains understudied. Here, we propose and evaluate a lightweight top-down attention mechanism for efficiently selecting modalities inside a global workspace. We show that our attention mechanism improves robustness to structured modality corruptions on two multimodal datasets of increasing complexity: Simple Shapes and MM-IMDb 1.0, and that the learned selection strategy generalizes across tasks, corruption regimes, and even modalities; capabilities not shared by some multimodal attention models from the literature. Beyond explicit corruption settings, on the MM-IMDb 1.0 benchmark, we show that the same mechanism improves the global workspace over its no-attention counterpart and yields decent benchmark performance. We conclude that robustness can be partly delegated to a small modality-wise controller, yielding a favorable robustness/complexity trade-off for multimodal integration

Robust multimodal systems must remain effective when some modalities are noisy, degraded, or unreliable. Existing multimodal fusion methods often learn modality selection jointly with representation learning, making it difficult to determine whether robustness comes from the selector itself or from full end-to-end co-adaptation. Motivated by Global Workspace Theory (GWT), we study this question using a lightweight top-down modality selector operating on top of a frozen multimodal global workspace. We evaluate our method on two multimodal datasets of increasing complexity: Simple Shapes and MM-IMDb 1.0, under structured modality corruptions. The selector improves robustness while using far fewer trainable parameters than end-to-end attention baselines, and the learned selection strategy transfers better across downstream tasks, corruption regimes, and even to a previously unseen modality. Even without explicit corruption, on MM-IMDb 1.0, we show that the same mechanism improves the global workspace over its no-attention counterpart and yields performance on par with the literature.

  \and Global Workspace
\end{abstract}

\section{Introduction}

In Multimodal machine learning, robustness to missing or degraded modalities is a central open problem\cite{baltrusaitis2019multimodal,wu2024mlmm_survey,qiu2024benchmarking}: will the trained systems work in real-world scenarios, where sensors may fail and data may be corrupted or even missing~\cite{ngiam2011multimodal,neverova2016moddrop,ma2021smil}? In practice, to solve the robustness problem, one is often forced to choose between scaling large end-to-end systems that demand vast amounts of data, training, and deployment budgets \cite{strubell2019energy,bommasani2021foundation}, or using smaller fusion mechanisms (e.g., gating- and attention-based fusion) that must be retrained whenever the downstream task or available modalities change. This raises a more specific question: can robustness be delegated to a lightweight modality selector, rather than relying on full end-to-end co-adaptation of the entire multimodal system?

Existing approaches span a range of fusion mechanisms: gated units such as GMU use multiplicative gates to modulate each modality’s contribution to a shared representation \cite{arevalo2017gmu}; dynamic methods such as DynMM use a gating function to select between uni- and multimodal experts (or fusion operations), yielding data-dependent computation paths \cite{xue2023dynmm}. Such models typically learn representations and fusion jointly in end-to-end systems. Thus, it is unclear how much robustness is due to the controller itself, versus the representations it co-adapts with.

Global Workspace Theory (GWT) \cite{baars1988cognitive,dehaene1998neuronal,dehaene2011gnw} offers a natural architectural perspective on this question. GWT proposes that specialized brain systems (vision, language, motor control, memory, etc.) coordinate via a shared, capacity-limited “global workspace”. Information is translated from the specialist modules into an amodal workspace representation, and an attentional mechanism acts as a spotlight that selects which source(s) can write to it before the result is broadcasted back to the specialists.

Previous multimodal systems have explored various aspects of GWT ~\cite{bao2020gwn,goyal2021sgw,sun2025ait,dossa2024gwagent}, but none isolate the attention mechanism totally : they also generally entangle attention with representation learning in end-to-end networks. We hypothesize that a small controller on top of a frozen multimodal workspace could flexibly re-weight modalities under changing reliability conditions, without retraining the whole system; this architecture, while computationally simple, remains compatible with GWT~\cite{vanrullen2021tins,devillers2025gwsemi,Maytie2024ZeroShotGW,Maytie2025MultimodalDreaming}.

We study this hypothesis in a controlled setting based on static, paired multimodal data, rather than dynamic input sequences such as video streams. This ensures changes in performance can be attributed to the attention mechanism itself, rather than to sequence modeling or cross-time alignment.

\begin{figure}[t]
  \centering
  \includegraphics[width=\textwidth]{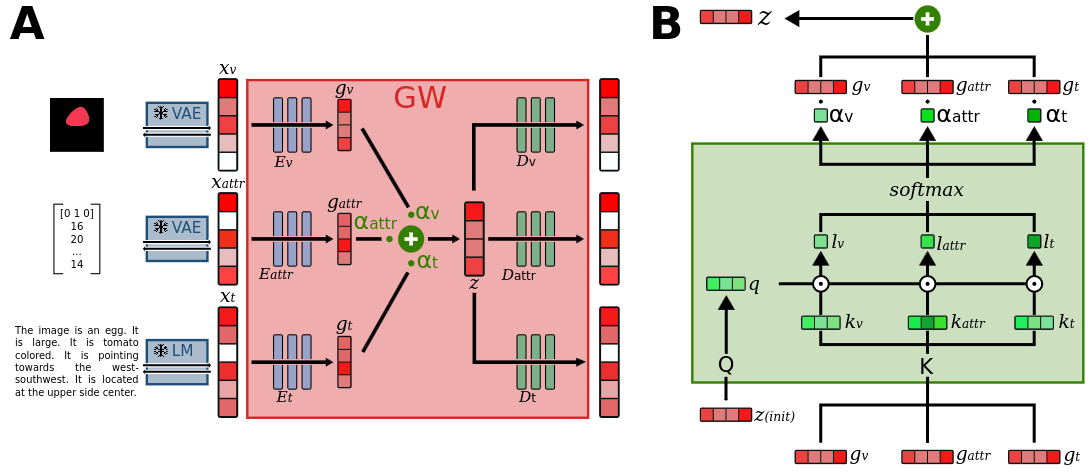}
  \caption{Architecture overview (illustrated on Simple Shapes).
\textbf{A}: Each modality is encoded by a frozen backbone, projected into the shared GW space by a modality-specific encoder $E_i$, fused into a single workspace latent $z$, and decoded back to modality-specific latent spaces by GW decoders $D_i$. During GW pretraining, fusion weights $\alpha_i$ are sampled at random.
\textbf{B}: The attention mechanism replaces random fusion with learned modality weights. It builds a query from an initial uniformly fused GW latent $z_{(init)}$ and keys from the modality-specific GW latents, then computes attention weights $\alpha_i$ by query-key similarity followed by a softmax. These weights define the final fused workspace latent $z$.}
  \label{fig:method}
\end{figure}

In this work, we build directly on the global latent workspace (GLW) framework of \cite{devillers2025gwsemi}, which learns an amodal latent space between pretrained, frozen modules. We slightly augment their method with a fusion mechanism. That is, on
top of the pretrained, frozen multimodal workspace, we introduce a small, modality-wise attention mechanism that:
\begin{itemize}
  \item yields higher corruption robustness than baselines from the literature on a $20\text{--}25\times$ smaller parameter budget
  \item transfers its learned selection strategy to unseen downstream tasks, and even to an unseen modality, without any finetuning
  \item when tested on the MM-IMDb benchmark without manual corruption, still improves performance over the non-attention baseline, and performs well in the context of recent models from the literature.
\end{itemize}

\section{Method}

Figure~\ref{fig:method}-A summarizes our architecture. We build directly on the global latent workspace (GLW) framework of~\cite{devillers2025gwsemi}, which learns an amodal shared space between pretrained, frozen modules using translation, demi-cycle, cycle-consistency, and contrastive alignment objectives. In the original formulation, each training step writes a \emph{single} modality into the workspace and decodes it into one or more target modalities. Our only modification at this stage is to allow the workspace to be written to by \emph{multiple modalities at once} through the fusion operator in Eq.~\eqref{eq:fuse}. This lets the same representation-learning objectives operate on fused multimodal inputs rather than only on one-to-one broadcasts.

\paragraph{Observed modality subsets}
Let $\mathcal{M}={M_1,\dots,M_n}$ be the set of modalities targeted for representation learning. 
For a given sample, we observe a subset $\mathcal{M}'\subseteq\mathcal{M}$. 
For any non-empty encoder subset $\mathcal{M}''\subseteq\mathcal{M}'$ we fuse the corresponding GW latents using Eq.~\eqref{eq:fuse}.

\paragraph{Encoding, fusion and decoding}
Each modality $M_i$ is first encoded by a pretrained and frozen backbone into a modality-specific latent vector $\mathbf{x}_i \in \mathbb{R}^{d_i}$. A learnable GW encoder $E_i:\mathbb{R}^{d_i}\rightarrow\mathbb{R}^{d}$ maps this latent into a GW representation
\begin{equation}
\mathbf{g}_i \;=\; E_i(\mathbf{x}_i)\in\mathbb{R}^d.
\end{equation}

For any non-empty encoder subset $\mathcal{M}''\subseteq\mathcal{M}'$, we fuse the corresponding GW latents into
\begin{equation}
\mathbf{z}_{\mathcal{M}''}
=
f\!\left(
\sum_{M_i\in\mathcal{M}''}\alpha_i \mathbf{g}_i
\right),
\label{eq:fuse}
\end{equation}
where $f(\cdot)=\tanh$ and $\boldsymbol{\alpha}=(\alpha_i)_{M_i\in\mathcal{M}''}$ is a simplex vector, i.e.\ $\alpha_i\ge 0$ and $\sum_i \alpha_i=1$.
This fused workspace state can then be decoded back into any target modality $M_j$ through the corresponding GW decoder $D_j$, yielding a prediction $\hat{\mathbf{x}}_j = D_j(\mathbf{z}_{\mathcal{M}''})$.

\subsection{Representation learning objectives}
\label{sec:repobj}

After decoding, losses can be computed exactly as in the GLW framework of~\cite{devillers2025gwsemi}, with the particular objective depending on the encoder subset $\mathcal{M}''$ and decoded target modality set.

Concretely, for each observed sample we enumerate non-empty encoder subsets $\mathcal{M}''\subseteq\mathcal{M}'$, fuse their GW latents into $\mathbf{z}_{\mathcal{M}''}$, and decode from this fused representation back to modality-specific latent spaces through the GW decoders. During this representation-learning stage, fusion weights can be sampled at random from a temperature-scaled softmax over i.i.d.\ scores; (in which case attention is not used when learning the workspace itself). Or, when useful, the attention mechanism can also be trained alongside the workspace. We always specify in our experiments which case is relevant.

As in~\cite{devillers2025gwsemi}, we optimize four complementary losses:  
(i) a \emph{demi-cycle} loss, which reconstructs the input modalities from the fused workspace state;  
(ii) a \emph{translation} loss, which predicts the other observed modalities from that same fused state;  
(iii) a \emph{cycle-consistency} loss, which broadcasts through complementary modalities and back to the original inputs; and  
(iv) a \emph{contrastive} loss, which aligns pre-fusion GW representations across modalities.

We optimize a weighted sum of translation, demi-cycle, cycle-consistency, and contrastive alignment losses. 

The full set-to-set loss formulation and explicit objective definitions are given in Supplementary Sec.~S1; since Devillers et al.~\cite{devillers2025gwsemi} already provide an extensive ablation study of these objectives, we do not repeat that analysis here.

\subsection{Top-down modality attention}
\label{sec:attn}

Figure~\ref{fig:method}-A summarizes our attention mechanism, which replaces the random fusion weights $\boldsymbol{\alpha}$ in Eq.~\eqref{eq:fuse} with \emph{data-dependent} modality weights.
It computes a query from a fused GW state, and compares it to modality-specific keys derived from the pre-fusion GW latents.

Given pre-fusion latents $\{\mathbf{g}_i\}_{M_i\in\mathcal{M}''}$, we use linear maps $K,Q:\mathbb{R}^d\rightarrow\mathbb{R}^h$ (with $K$ shared across modalities) to form
\[
\mathbf{k}_i = K(\mathbf{g}_i)\in\mathbb{R}^h,\qquad
\mathbf{q}=Q(\mathbf{z}^{\mathrm{init}})\in\mathbb{R}^h.
\]

Since the query depends on a fused GW state, we first initialize it with uniform weights:
\[
\alpha_i^{(\mathrm{init})}=\tfrac{1}{|\mathcal{M}''|},\qquad
\mathbf{z}^{(\mathrm{init})}=f\big(\sum_{M_i\in\mathcal{M}''}\alpha_i^{(\mathrm{init})}\mathbf{g}_i\big).
\]

And lastly, we score each modality by dot-product similarity to the query and normalize across modalities:
\[
\ell_i=\langle \mathbf{k}_i,\,\mathbf{q}\rangle,\qquad
\alpha_i=\operatorname{softmax}_{i\in\mathcal{M}''}(\ell_i)
\]

\section{Experimental setup}

\subsection{Datasets and training protocol}

\textit{For reproducibility, we provide full backbone choices, hidden sizes, loss coefficients, and training settings for all experiments in Supplementary Sec.~S2.}

We evaluate the proposed selector on two multimodal datasets of increasing complexity.

\textit{Simple Shapes} \cite{devillers2025gwsemi} is a synthetic dataset of geometric shapes (diamonds, triangles, eggs) with a given size, position, color, and rotation. They can be described in three modalities: text, images, and attribute vectors. We use the original large-scale dataset to pretrain the frozen modality backbones and the global workspace (GW). For downstream evaluation, we use a restricted classification version with five quantized attribute-prediction tasks (shape, color, rotation, position, and size), designed so that clean in-distribution classification is easy and performance differences primarily reflect robustness under corruption. Exact quantization rules and dataset sizes are reported in Supplementary Sec.~S2.1.

\textit{MM-IMDb} \cite{arevalo2017gmu} is an ideal choice for our experiments: movie synopses (text) and posters (image) are used to classify the genre of each movie (label). In explicit-corruption robustness tests, we input the image and text modalities. In the secondary clean-data experiment, we include multi-hot labels as a third input modality.

In both datasets, we encode all the samples through frozen pretrained backbones, and use these latent vectors as inputs to the GW (or baseline method) encoders.

\subsection{Training stages and corruption}\label{sec:reproducibility}

On Simple Shapes, inputs are corrupted through adding varying levels of Gaussian noise to latent inputs. On MM-IMDb, we use the image and text corruption taxonomy of \cite{qiu2024benchmarking} at maximum severity, and use between 0 and 4 corruptions on top of each other, uniformly and independently chosen, to control task difficulty. 

For all explicit-corruption robustness tests (both MM-IMDb and Simple Shapes), our experiments follow a multi-stage protocol. First, we train the GW representation model on clean data using random fusion weights.
Second, with the GW frozen, we train downstream classification probes on GW fused latents on clean data.
Third, we add corruption to two out of three modalities on Simple Shapes, and one out of two modalities on MM-IMDb (the "clean" modality, and level of corruption being random per train sample). With this corruption, we
train the modality selector with the same loss as the probes, while keeping both the GW and the probes frozen,
so that any robustness gains can be attributed to modality selection alone.

For the secondary clean MM-IMDb experiment, we instead train the selector jointly with the GW to facilitate multimodal integration in the absence of explicit corruption. 
 In this specific case, we use CLIP and BLIP-2 backbones following \cite{dufumier2025comm}. Full architecture and optimization details are reported in Supplementary Sec.~S2.

\section{Results}

\subsection{Noise Robustness on Simple Shapes and MM-IMDb}

We follow multi-stage training as defined in \ref{sec:reproducibility}.

\begin{figure}[t]
\centering
\includegraphics[width=0.8\linewidth]{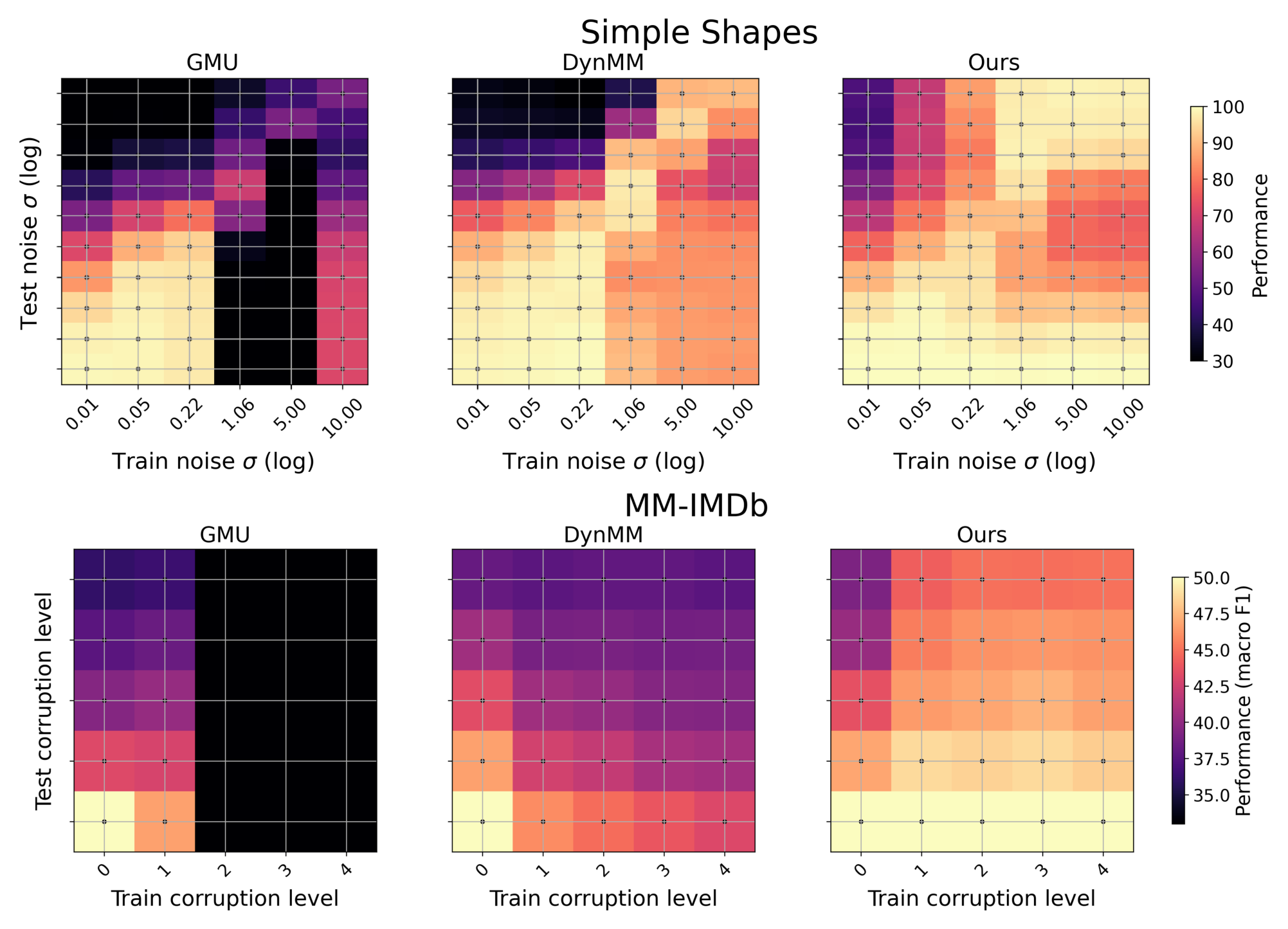}
\caption{Accuracy / macro-F1 heatmaps across train and test corruption levels for Simple Shapes and MM-IMDb.}
\label{fig:robustness_pair}
\vspace{-1.em}
\end{figure}

First, we aim to show that our attentional system can learn to select modalities so as to improve robustness on a downstream task. We note that in our setting, maximum noise on Simple Shapes leads to chance level performance (23 percent), and on MM-IMDb, 4 corruptions being applied leads to a 7 percent drop in performance. This allows us to test the attention mechanism on both strong, artificial noise, and weaker, more realistic corruptions.

As baselines, we use GMU \cite{arevalo2017gmu} and DynMM \cite{xue2023dynmm}. We choose GMU (a classic feature-wise attention mechanism) to assess whether feature-wise attention could surpass our method by tuning out the features corresponding to the noise. DynMM applies a gating (soft during training, hard during testing) mechanism between multiple possible fusion methods: average, max, concatenation followed by a linear layer, and choice of either modality alone. We use it as a baseline because its attention mechanism can learn to use a single score per modality if it chooses the single-modality options. For a fair comparison, we use the same MLP sizes and fused latent sizes as in our GW architecture. Both baselines are finetuned end-to-end on the classification tasks, with noised inputs, making them much heavier than ours. On Simple Shapes, only 4,544 selector parameters are trained on the corruption task, versus 92,578 for DynMM and 113,340 for GMU; on MM-IMDb, we use 0.06M parameters for corruption tuning-out, versus 4.48M and 5.01M for DynMM and GMU, respectively. Full parameter-count breakdowns are reported in Supplementary Table~S1.

In Fig.~\ref{fig:robustness_pair}, we observe that all models struggle when trained on a certain corruption level and evaluated on a different level. Still, although our probes and Global Workspace encoders have never encountered any corruption, our model holds up well for in-distribution levels of noise, and outperforms the other models for most out-of distribution noise levels. On Simple Shapes, for instance, our attention system trained with only $\approx4,000$ parameters can better adapt to new input statistics than dedicated attention systems trained with 20--25 times more parameters. After training at maximum corruption, we found that attention scores to corrupted modalities systematically decrease with corruption level: MM-IMDb (Pearson R = -0.2946, p< 1e-100) and Simple Shapes (R = -0.6432, p<1e-100). In the next section, we explore whether the attention mechanism can also generalize to unseen classification tasks, or unseen modalities.

\subsection{Generalization across tasks and modalities}

\subsubsection{Leave-out task}

We argue that one of the advantages of our GW attention is more flexible processing. This could manifest as the ability to learn behavior on one task, and transfer it to the next without retraining attention parameters.
On Simple Shapes, to put this to the test, we start from a pretrained GW and 5 classification probes trained on clean data, just as in the previous experiment. We then introduce the same noise schedule to train the attention mechanism, but only train on one of the 5 tasks.
At inference, we test this attention mechanism on the 4 other (out-of-distribution) probes. In all the following generalization tests, we set the noise standard deviation to $\sigma=5$ for both training and evaluation, a high level ensuring that the task can only be solved by tuning out the noise.

We also test our hypothesis on the more realistic MM-IMDb with our realistic corruptions. We divide the target labels into 4 arbitrary subsets; predicting one subset of labels with the pretrained probe corresponds to a \textit{task}. We use the maximum level of corruption (4). As before, we train our attention mechanism on one task, and test it on the 3 other probes.

As baselines, we train DynMM and GMU systems end-to-end on a single classification task, freeze their encoders (which are the same size as ours) and fusion mechanisms, and train new probes for the left-out tasks, using the frozen fused representation (same size as ours). Additionally, we also measure the performance of our GW (and classification probes) in the same conditions but without attention, i.e. with random fusion scores.

\begin{figure}[t]
\centering
\includegraphics[width=\linewidth]{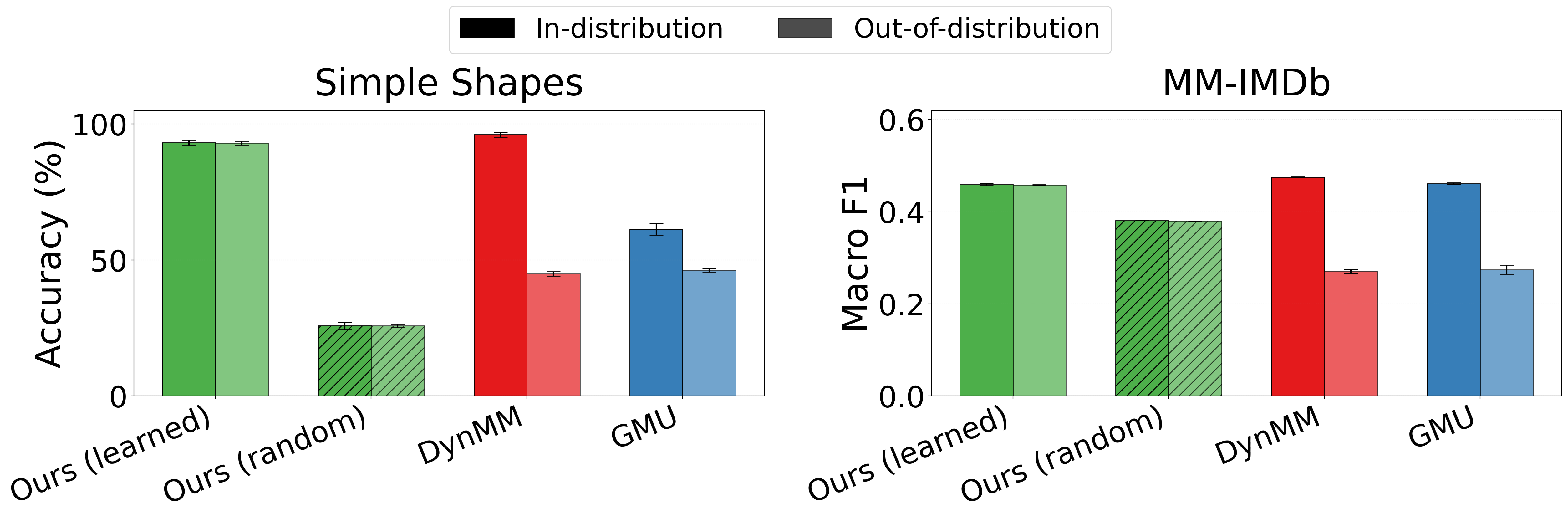}
\caption{Leave-out task generalization on Simple Shapes and MM-IMDb, comparing in-distribution and out-of-distribution evaluation across tasks.}
\label{fig:leave_out_task_generalization}
\vspace{-1.2em}
\end{figure}

In Fig.~\ref{fig:leave_out_task_generalization}, we see that our GW attention generalizes to the left-out tasks, reliably removing corruption from the learned representation in our multi-stage training paradigm. The attention mechanism has never seen the left-out tasks, and the probes and GW have never seen the corruption; nonetheless, on both datasets and types of noise, attention provides equal improvement over the random baseline in in-distribution as in out-of-distribution settings. This means it is able to use one task scenario to learn the signatures of reliable vs. unreliable modalities, in a way that can generalize to other tasks scenarios. In contrast, other attention systems (DynMM and GMU) learn task- and corruption-specific strategies that do not transfer well to out-of-distribution settings. Having validated that our representation and attention methods can transfer knowledge from task to task, we now ask if they can learn a strategy for one modality and apply it to another.

\subsubsection{Modality-wise generalization: unseen clean modality}

Can our attention mechanism apply knowledge learned on two modalities to the third one? To test this, we change the noise schedule so that one of the three modalities is never provided as clean during training (as before, only one modality can be clean at a time). We expect the trained attention system to assign a low or negligible weight to this modality. At inference, we go back to the normal noising schedule where any modality can be clean; will the trained attention system be able to assign a high weight to the previously left-out modality when it is the only clean one? As this test requires at least 3 modalities, it can only be performed on Simple Shapes.

As baselines, we include GMU and DynMM (with the same encoder and fused latent sizes as our method), training them end-to-end on the new restricted noise schedule, and evaluating on the normal noise schedule. Since these two methods normally train their probes in the final noisy conditions, we finetune the classification probes (but not the rest of the attention systems) on the normal noise schedule before testing. As always, we also include the GW random fusion baseline, i.e. our model without attention.

\begin{figure}[t]
\centering
\includegraphics[width=0.6\linewidth]{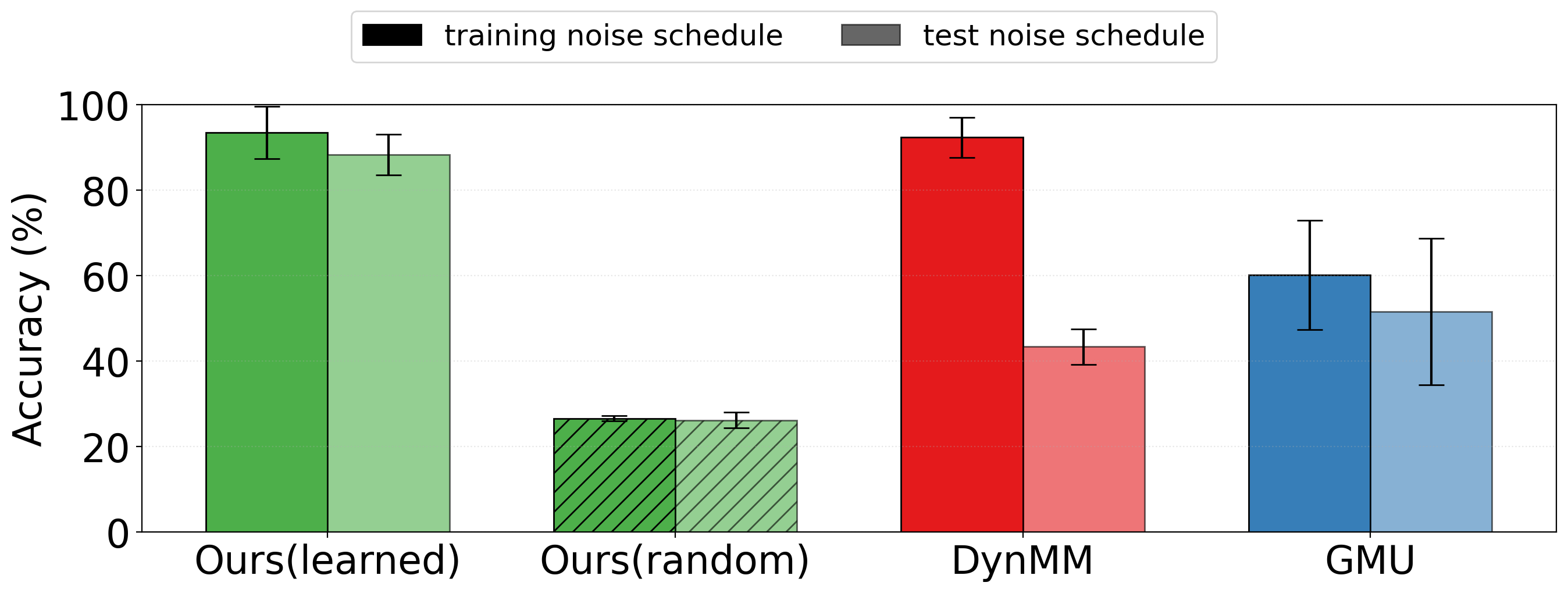}
\caption{Modality generalization test (unseen clean modality). One modality is always noised during training and may be clean at test time. Results are averaged across left-out modalities and tasks; GMU and DynMM probes are fine-tuned before testing, while our model is not.}
\label{fig:modality_generalization_unseen_clean}
%\vspace{-1.em}
\end{figure}

In Fig.~\ref{fig:modality_generalization_unseen_clean}, we see that for our model, removing the constraint on the ``always-noised'' modality at inference time results in less than 5 percent accuracy drop. In contrast, performance for the random fusion baseline collapses. Thus, our system was able to learn efficient key and query matrix weights from one modality, and apply them to another without retraining.

On the contrary, for GMU and DynMM baselines, performance is much lower when changing the noise schedule, because they are trained end-to-end on one configuration and struggle to transfer their learned strategy to an unseen configuration.
In this test, we aimed to leave out one modality from attention training by always noising it; in Supplementary Sec.~3.2 we show similar properties of our system with a modality totally left-out during training, alongside an analysis of the actual attention weights output by the system.

\subsection{Utility beyond explicit corruption on MM-IMDb}
\label{sec:beyond-explicit}
\begin{table}[t]
\caption{MM-IMDb 1.0 macro-F1. For CoMM, we draw performance values from \cite{dufumier2025comm}. For GMU and BridgeTow, we use values from \cite{Li2023IDKG}. By ``end-to-end'' we mean the entire model was trained or finetuned directly on the final classification objective.}
\label{tab:mmimdb_results_full}
\centering
\scriptsize
\setlength{\tabcolsep}{2pt}
\renewcommand{\arraystretch}{1.12}
\begin{tabularx}{\columnwidth}{@{}X c c c@{}}
\toprule
\textbf{Model} & \textbf{Macro-F1} & \textbf{End-to-end} & \textbf{Train FLOPs (est.)} \\
\midrule
GMU (Arevalo et al., 2017)                      & 54.10          & yes & -- \\
CoMM (CLIP, finetuned)                          & 58.97$\pm$0.19 & yes & -- \\
CoMM (BLIP-2, finetuned)                        & 62.00$\pm$0.25 & yes & $> 4.7 \cdot 10^{15}$ \\
BridgeTow \cite{huang2023kosmos}                & 63.30          & --  & -- \\
\addlinespace[2pt]
\cmidrule(lr){1-4}
\addlinespace[2pt]
Ours (GW, CLIP, random)                         & 50.00$\pm$1.99 & no  & -- \\
Ours (GW, CLIP)                                 & 53.48$\pm$0.37 & no  & -- \\
CoMM (CLIP, linear readout) \cite{dufumier2025comm} & 54.63$\pm$0.22 & no  & -- \\
CoMM (BLIP-2, linear readout)                   & 58.44$\pm$0.43 & no  & $4.7 \cdot 10^{15}$ \\
Ours (GW, BLIP-2, random)                       & 60.41$\pm$0.52 & no  & $7.5 \cdot 10^{14}$ \\
Ours (GW, BLIP-2)                               & 65.34$\pm$0.14 & no  & $7.5 \cdot 10^{14}$ \\
\bottomrule
\end{tabularx}
\vspace{-1.em}
\end{table}

Our attention mechanism proved helpful in terms of robustness to corrupted modalities and generalization to out-of-distribution conditions. As a secondary validation, we now want to see if it remains useful in a clean multimodal setting, where informative modalities are defined by potentially more subtle cues than the presence or absence of corruptions. We thus use MM-IMDb 1.0 without any corruption for this purpose. 
To contextualize our performance, we include multiple baselines from the literature, ranging from representation learning methods \cite{radford2021clip,dufumier2025comm,huang2023kosmos}, to task-oriented, end-to-end methods \cite{arevalo2017gmu,baltrusaitis2019multimodal}.
As an ablation to verify that our attention mechanism is actually useful, we also include the random attention scores version of our method.

Results, alongside efficiency metrics of our model and training procedure are reported in Table~\ref{tab:mmimdb_results_full}. We find that our model is efficient when compared with CoMM, requiring significantly less floating-point operations to train, while using approximately the same number of parameters.

These results should be interpreted carefully.
In our MM-IMDb setting, labels are used as a training-time modality for workspace representation learning (and of course never provided as inputs at test time), which is not the case for CoMM and BridgeTow. Meanwhile, the highest performing methods are finetuned end-to-end specifically for the final classification task, whereas our global workspace is not.
What's more, we average-pool backbone tokens to enter the MLPs, which CoMM does not require. This also explains the performance gap between our CLIP and BLIP-2 settings, with 512-dim and 2048-dim inputs to the MLPs respectively. We note that the GW and CoMM both have low-dimension fused bottlenecks (512 and 768 dimensions respectively), so differences in performance do not arise from lack of compression in CoMM.
We therefore do not interpret Table~\ref{tab:mmimdb_results_full} as a strict apples-to-apples ranking. Rather, we draw two limited conclusions: first, the learned-attention variant consistently improves over the corresponding random-fusion baseline, even though there is no explicit corruption. Second, absolute levels of performance are on par with the state-of-the-art, with a favorable performance-to-training-expense ratio.

\section{Discussion}

GWT provides a framework for flexible multimodal integration in AI systems. Prior work has revealed the potential of this framework for downstream classification tasks~\cite{devillers2025gwsemi} or for cross-modal transfer in RL agents~\cite{Maytie2024ZeroShotGW,Maytie2025MultimodalDreaming}. In this work, we proposed a possible implementation for an attention controller in a Global Workspace architecture, and explored its properties. We complement the architecture of \cite{devillers2025gwsemi} with a fusion operator, and an attention mechanism that intelligently selects the input modalities; depending on the application, this attention mechanism can be trained along with the global workspace for representation learning objectives, or separately for a specific downstream task.

On a synthetic dataset of geometric shapes and on the more naturalistic MM-IMDb 1.0 dataset, we demonstrated that training the attention mechanism with a frozen GW representation makes the system more robust to corruption than baselines from the literature. The attention system focused the multimodal fusion weights on the clean modality and ignored the corrupted one(s). We also found that our system could generalize to new, out-of-distribution conditions, whether of corruption patterns, of unseen downstream tasks, or even new modalities, without any finetuning.

We also evaluated our system on MM-IMDb outside the explicit corruption regime used in our main experiments, and reported a favorable performance-to-efficiency ratio. The learned selector still improved performance over the no-attention baseline, suggesting that its utility extends beyond perturbation-focused problems: for some movies, more attention was focused on the poster, for others on the text synopsis, and as a result, the GW was better at predicting the movie genre label.

Some limitations remain to be addressed in future work.
Dynamical processing could thus be the logical next goal for our GW system, because our attention mechanism (Figure~\ref{fig:method}B) can in principle operate iteratively over multiple time steps. Second, the architecture presented in this work could, in theory, scale to any number of modalities (sound, infrared sensors, etc.). This is another promising future avenue.

\section*{Acknowledgements}
This work was supported by ANITI (ANR-19-P3IA-0004, ANR-23-IACL-0002), ANR COCOBOT (ANR-21-FAI2-0005), Région Occitanie through ``Défi Clé Robotique centrée sur l'humain'', and the European Union/ERC Advanced GLOW project (101096017).

% ---- Bibliography ----
% BibTeX users should specify bibliography style 'splncs04'.
\bibliographystyle{splncs04}
\bibliography{icann_2026}

% ==================================================
% Supplementary material
% ==================================================

\clearpage

% Reset counters for supplement numbering.
\setcounter{section}{0}
\setcounter{subsection}{0}
\setcounter{subsubsection}{0}
\setcounter{figure}{0}
\setcounter{table}{0}
\setcounter{equation}{0}
\setcounter{footnote}{0}

% Supplement numbering: S1, Fig. S1, Table S1, Eq. S1.
\renewcommand{\thesection}{S\arabic{section}}
\renewcommand{\thesubsection}{S\arabic{section}.\arabic{subsection}}
\renewcommand{\thesubsubsection}{S\arabic{section}.\arabic{subsection}.\arabic{subsubsection}}
\renewcommand{\thefigure}{S\arabic{figure}}
\renewcommand{\thetable}{S\arabic{table}}
\renewcommand{\theequation}{S\arabic{equation}}

% Avoid duplicate label collision:
% main.tex and supp.tex both define \label{sec:reproducibility}.
\makeatletter
\let\origlabel\label
\def\duplicatelabel{sec:reproducibility}
\def\label#1{%
  \def\currentlabel{#1}%
  \ifx\currentlabel\duplicatelabel
    \origlabel{sec:supp_reproducibility}%
  \else
    \origlabel{#1}%
  \fi
}
\makeatother

\title{Supplementary Material for\\
An Attention Mechanism for Robust Multimodal Integration in a Global Workspace Architecture}

\titlerunning{Supplementary Material}

\author{
Roland Bertin-Johannet\inst{1}\orcidID{0009-0007-4883-2999}\thanks{Corresponding author: \email{[roland.bertin-johannet@proton.me]}}
\and Lara Scipio\inst{1}
\and Leopold Maytié\inst{1}
\and Rufin VanRullen\inst{1}
}

\institute{
CerCo, CNRS, Université de Toulouse; ANITI, Artificial and Natural Intelligence Toulouse Institute, Toulouse, France
}
\maketitle

\noindent
This supplementary document provides extended method details, full reproducibility information, and additional experimental results that could not be included in the main paper due to space constraints. All code used in this work is publicly available at:\newline
\url{https://github.com/RolandBERTINJOHANNET/GW_attention}.

% ==================================================
\section{Extended Method Details}
\label{sec:supp_method}
% ==================================================
\subsection{Notation and modality sets}
\label{subsec:supp_sets}

For consistency with the main paper, we use the same notation throughout this supplementary section. Each modality $M_i$ is first encoded by a pretrained and frozen backbone into a modality-specific latent vector $\mathbf{x}_i\in\mathbb{R}^{d_i}$. A learnable GW encoder $E_i:\mathbb{R}^{d_i}\rightarrow\mathbb{R}^{d}$ maps this latent to a GW representation
\begin{equation}
\mathbf{g}_i \;=\; E_i(\mathbf{x}_i)\in\mathbb{R}^{d}.
\label{eq:supp_gi}
\end{equation}

Let $\mathcal{M}=\{M_1,\dots,M_n\}$ denote the full set of modalities targeted for representation learning. Because samples may contain different subsets of modalities, for a given training sample we observe only a subset $\mathcal{M}'\subseteq\mathcal{M}$. At training time, we consider all non-empty encoder subsets $\mathcal{M}''\subseteq\mathcal{M}'$ when forming fused GW representations.

For any non-empty encoder subset $\mathcal{M}''\subseteq\mathcal{M}'$, we define the fused GW latent
\begin{equation}
\mathbf{z}_{\mathcal{M}''}
\;=\;
f\!\left(\sum_{M_i\in\mathcal{M}''}\alpha_i\,\mathbf{g}_i\right),
\label{eq:supp_fuse}
\end{equation}
where $f(\cdot)=\tanh$ and $\boldsymbol{\alpha}=(\alpha_i)_{M_i\in\mathcal{M}''}$ is a simplex vector satisfying $\alpha_i\ge 0$ and $\sum_i\alpha_i=1$.

When defining the full representation-learning objectives, it is also convenient to introduce a generic non-empty \emph{input} subset $\mathcal{I}\subseteq\mathcal{M}'$ and a non-empty \emph{output} subset $\mathcal{O}\subseteq\mathcal{M}'$. We also use $\overline{\mathcal{M}''}=\mathcal{M}\setminus\mathcal{M}''$ to denote the complementary modality set. This complement may include modalities that are not observed for a given sample; in that case, the cycle-consistency objective uses predicted latents rather than direct supervision on these modalities.

\subsection{Broadcast operator}
\label{subsec:supp_broadcast}

Following the spirit of~\cite{devillers2025gwsemi}, we define a single set-to-set broadcast primitive and derive the representation-learning objectives from it. Given a non-empty input subset $\mathcal{I}\subseteq\mathcal{M}'$ and a non-empty output subset $\mathcal{O}\subseteq\mathcal{M}'$, we define
\begin{equation}
\operatorname{Br}_{\mathcal{I}\rightarrow\mathcal{O}}
\big(\{\mathbf{x}_i\}_{i\in I}\big)
=
\left\{
D_j\!\left(
f\!\left(
\sum_{i\in I}\alpha_i E_i(\mathbf{x}_i)
\right)
\right)
\right\}_{j\in O},
\label{eq:supp_broadcast}
\end{equation}

where $I$ and $O$ denote the index sets corresponding to modality subsets $\mathcal{I}$ and $\mathcal{O}$, and where $D_j$ and $E_i$ are the learnable GW decoder and encoder associated with modalities $M_j$ and $M_i$.

This formulation generalizes the original one-to-one GLW write operations of~\cite{devillers2025gwsemi}. In the original setup, the workspace is written to by a single modality at a time. Here, the same broadcast principle is retained, but the write operation can originate from a fused subset of modalities through Eq.~\eqref{eq:supp_fuse}. This is the only representational change we introduce before adding the top-down selector.

\subsection{Random fusion during representation learning}
\label{subsec:supp_random_fusion}

Unless stated otherwise, all representation-learning objectives use random fusion weights whenever Eq.~\eqref{eq:supp_fuse} is invoked. For a given non-empty encoder subset $\mathcal{M}''\subseteq\mathcal{M}'$, we sample one i.i.d.\ score $s_i\sim\mathcal{U}(0,1)$ for each $M_i\in\mathcal{M}''$, and define
\begin{equation}
\alpha_i
=
\frac{\exp(s_i/\tau)}
{\sum_{M_k\in\mathcal{M}''}\exp(s_k/\tau)},
\qquad M_i\in\mathcal{M}'',
\label{eq:supp_random_fusion}
\end{equation}
with temperature $\tau>0$.

Thus, during representation learning, the workspace is trained under varying fused input configurations, but the learned top-down attention mechanism is not yet used. In the main paper, this point is important because the later robustness gains can then be attributed to the selector itself rather than to co-adaptation of the selector with the representation-learning stage.

\subsection{Full representation-learning objectives}
\label{subsec:supp_losses}

We build directly on the global latent workspace framework of~\cite{devillers2025gwsemi}, which optimizes translation, demi-cycle (reconstruction), cycle-consistency, and contrastive alignment between pretrained modules through an amodal shared space. Devillers et al.~\cite{devillers2025gwsemi} showed through systematic ablations that these components play complementary roles: self-supervised objectives such as cycle- and demi-cycle-consistency help reduce dependence on paired multimodal supervision, while translation-style broadcasts improve generalization relative to purely contrastive alignment.

Our key modification is that the workspace can now be written to by multiple modalities at once through Eq.~\eqref{eq:supp_fuse}. We therefore generalize their one-to-one formulation by expressing the objectives as instances of the set-to-set broadcast primitive $\operatorname{Br}_{\mathcal{I}\rightarrow\mathcal{O}}$ in Eq.~\eqref{eq:supp_broadcast}. In our implementation, for each encoder subset we decode to all modalities in $\mathcal{M}$; demi-cycle and translation losses are defined on observed subsets of $\mathcal{M}'$, while cycle-consistency uses the complementary modality set $\overline{\mathcal{M}''}=\mathcal{M}\setminus\mathcal{M}''$.

Let $\mathcal{L}_j(\cdot,\cdot)$ denote the modality-specific reconstruction or prediction loss for modality $M_j$ (In this work, we use MSE for continuous latents, and BCE for labels when we include a label modality in section \ref{sec:beyond-explicit}), with targets expressed in the corresponding backbone latent space.

\subsubsection{Demi-cycle objective}
\label{subsubsec:supp_dcy}

For any non-empty input subset $\mathcal{M}''\subseteq\mathcal{M}'$, the demi-cycle objective corresponds to reconstructing the modalities used to write to the workspace:
\begin{equation}
\mathcal{L}_{\text{dcy}}^{\mathcal{M}''}
\;=\;
\sum_{M_j\in\mathcal{M}''}
\mathcal{L}_j\Big(
\operatorname{Br}_{\mathcal{M}''\rightarrow\{M_j\}}
\big(\{\mathbf{x}_i\}_{M_i\in\mathcal{M}''}\big),
\;\mathbf{x}_j
\Big).
\label{eq:supp_dcy}
\end{equation}

This is the self-target special case of the broadcast primitive: the fused workspace state is required to preserve sufficient information to reconstruct the modalities that produced it.

\subsubsection{Translation objective}
\label{subsubsec:supp_tr}

Using the same encoder subset $\mathcal{M}''$, the translation objective broadcasts to the other observed modalities:
\begin{equation}
\mathcal{L}_{\text{tr}}^{\mathcal{M}''}
\;=\;
\sum_{M_j\in\mathcal{M}'\setminus\mathcal{M}''}
\mathcal{L}_j\Big(
\operatorname{Br}_{\mathcal{M}''\rightarrow\{M_j\}}
\big(\{\mathbf{x}_i\}_{M_i\in\mathcal{M}''}\big),
\;\mathbf{x}_j
\Big).
\label{eq:supp_tr}
\end{equation}

This encourages the fused workspace representation to remain predictive of the other modalities available for that sample.

\subsubsection{Cycle-consistency objective}
\label{subsubsec:supp_cycle}

We use a cycle-consistency loss as an unsupervised regularizer inspired by back-translation~\cite{Artetxe2017UnsupervisedNMT,Lample2018PhraseBasedNeuralUMT}, adapted to the set-to-set broadcast primitive.

 We first broadcast from the inputs $\mathcal{M}''$ to the complementary modalities:
\begin{equation}
\big\{\widehat{\mathbf{x}}_j\big\}_{M_j\in\overline{\mathcal{M}''}}
\;=\;
\operatorname{Br}_{\mathcal{M}''\rightarrow\overline{\mathcal{M}''}}
\big(\{\mathbf{x}_i\}_{M_i\in\mathcal{M}''}\big).
\label{eq:supp_cycle_forward}
\end{equation}

These predicted complementary latents are then treated as inputs and broadcast back to the original subset:
\begin{equation}
\big\{\widetilde{\mathbf{x}}_i\big\}_{M_i\in\mathcal{M}''}
\;=\;
\operatorname{Br}_{\overline{\mathcal{M}''}\rightarrow\mathcal{M}''}
\big(\{\widehat{\mathbf{x}}_j\}_{M_j\in\overline{\mathcal{M}''}}\big).
\label{eq:supp_cycle_backward}
\end{equation}

The cycle-consistency loss compares these reconstructions to the original inputs:
\begin{equation}
\mathcal{L}_{\text{cycle}}^{\mathcal{M}''}
\;=\;
\sum_{M_i\in\mathcal{M}''}
\mathcal{L}_i\big(\widetilde{\mathbf{x}}_i,\;\mathbf{x}_i\big).
\label{eq:supp_cycle}
\end{equation}

As in prior back-translation-style objectives, this term regularizes the shared space even when the forward pass traverses predicted intermediate modalities rather than observed ones.

\subsubsection{Contrastive alignment objective}
\label{subsubsec:supp_contrast}

In addition to reconstruction- and translation-based objectives, we align prefusion GW representations across observed modalities with a pairwise contrastive loss $\mathcal{L}_{\text{contrast}}$ (InfoNCE), following~\cite{devillers2025gwsemi}. Concretely, for modalities observed in the same sample, their encoded GW latents are treated as positive pairs, while the minibatch provides negatives. In this work, we keep this term unchanged relative to the original GLW framework; only the fused write operation differs.

\subsubsection{Overall objective}
\label{subsubsec:supp_lrep}

The final representation-learning objective is the weighted sum
\begin{equation}
\mathcal{L}_{\text{rep}}
=
\lambda_{\text{tr}}\,\mathcal{L}_{\text{tr}}
+
\lambda_{\text{dcy}}\,\mathcal{L}_{\text{dcy}}
+
\lambda_{\text{cycle}}\,\mathcal{L}_{\text{cycle}}
+
\lambda_{\text{contrast}}\,\mathcal{L}_{\text{contrast}},
\label{eq:supp_lrep}
\end{equation}
with $\lambda_{\text{tr}},\lambda_{\text{dcy}},\lambda_{\text{cycle}},\lambda_{\text{contrast}}\ge 0$.

A thorough ablation study of the effects of these loss components has already been reported in~\cite{devillers2025gwsemi}. Since our intervention at this stage is limited to replacing one-to-one writes by fused many-to-many writes, and our ultimate goal is to test the usefulness of modality-specific attention-scores in the fusion operation, we do not repeat that ablation study here.

% ==================================================
\section{Extended Experimental Setup and Reproducibility}
\label{sec:supp_setup}
% ==================================================

\subsection{Datasets}

\subsubsection{Simple Shapes}

First, we evaluate our approach on a synthetic dataset from \cite{devillers2025gwsemi}, the Simple Shapes Dataset. Samples from this dataset are geometric shapes (diamonds, triangles, eggs) with a given size, position, color, and rotation. They can be described in three modalities: text, images, and attribute vectors (the latter being a complete numerical description of the sample). The dataset comprises 1 million samples at train time, 50,000 for validation and 1000 at test time. This version of the dataset is used to pretrain the modality-specific backbones and the GW multimodal representation with the representation learning objectives described in Supplementary Sec.~S1.

We define 5 downstream classification tasks, one for each of the 5 attributes in the Simple Shapes Dataset. Only one of the four attributes is categorical so we thus build a second, restricted dataset specifically for training and testing classification heads. In it, we only use 9 different RGB values. For rotation, we separate the circle into 16 bins of width $\pi/8$, and only keep samples inside the 4 bins corresponding to top, bottom, left and right. For the position attribute, we split the 2D space into $7\times7$ bins (integer x and y positions between $[-3,3]$), and keep only samples inside four chosen bins: bottom $(x,y)=(0,-2)$, top $(0,2)$, left $(-2,0)$ and right $(2,0)$. Lastly, for size, we use 4 non-overlapping bins that cover the entirety of possible sizes. This makes the classification tasks intrinsically easy (for in-distribution, clean data samples), allowing us to focus our experiments on noise-robustness. For this classification dataset, we have 500,000 train samples, and 1,000 for testing.

\subsubsection{MM-IMDb 1.0}

The second dataset we use is MM-IMDb 1.0 \cite{arevalo2017gmu}. It is an ideal choice for our experiments: movie synopses (text) and posters (image) are used to classify the genre of each movie (label). The text and image sometimes provide redundant but sometimes conflicting information, and are not always equally informative. The dataset consists of 15,552 train movies, 2,608 for validation, and 7,799 for test, each represented by its synopsis (text) and poster (image). We use binary labels (across 23 distinct but not mutually exclusive genres), both as a third input modality (when training our GW architecture for multimodal representation objectives), and as a classification target (when training the attention selection mechanism; in this case, labels are never provided as inputs).

In both datasets, we encode all the samples through frozen pretrained backbones (acting as the ``modules'' of our GW architecture), and use these latent vectors as our GW inputs.

\subsection{Reproducibility details}
\label{sec:reproducibility}

\subsubsection{Simple Shapes}

On Simple-Shapes, we use \textit{simple-shapes-pretrained} VAEs as frozen backbones for images and attributes (latent sizes 8 and 10 respectively), and a pretrained RNN for text (one latent vector of size 64 per sample).
For the Global Workspace (GW), we use a latent dimension $d=32$ and implement each GW encoder/decoder as an MLP with two hidden layers of size 64.
Classification probes are 4-layer MLPs with hidden size 512, batch normalization, dropout $p=0.2$, and GELU activations.
Our modality attention uses linear Key and Query projections with head size $h=64$.
We use a batch size of 512.
Training: The GW is trained for 100{,}000 steps with Adam and a OneCycleLR schedule, using loss coefficients $\lambda_{\text{tr}}=\lambda_{\text{cycle}}=\lambda_{\text{dcy}}=1$ and $\lambda_{\text{contrast}}=0.01$. The GW is then frozen, while classification probes are trained for 3 epochs. Finally, with frozen probes, attention is trained for 5 epochs.

\subsubsection{MM-IMDb 1.0}

On MM-IMDb, we treat the labels (movie genres) as a third modality during GW multimodal representation learning.
GW size is 512, GW encoders and decoders are 4-hidden-layer MLPs of width 384, with residual connections skipping the first two hidden layers and skipping the last two hidden layers.
Our modality attention uses single-head linear Key and Query projections with head size $h=256$. We use the same loss coefficients as in the Simple Shapes case.

\paragraph{For the corruption robustness tests}
We pretrain the GW multimodal representation on labels, images and text without corruption. The GW is then frozen during classification probe training and attention training. We encode the synopsis text and poster images with \textit{BAAI/bge-base-en-v1.5} and \textit{facebook/dinov2-base} respectively, using the CLS token.

\paragraph{For the secondary, clean MM-IMDb experiment}
This time, we train our attention mechanism alongside the global workspace, to facilitate multimodal integration (rather than noise- or corruption-robustness).
We train a 3-modality Global Workspace in which labels are used as a modality during representation learning. Labels are never provided as inputs during evaluation; label prediction is obtained through the learned label decoder.
We adopt the same data augmentation scheme as \cite{dufumier2025comm}: default SimCLR augmentations for images and BERT-style token masking for text. We use the same backbones: CLIP (\textit{openai/clip-vit-base-patch32} for images; \textit{sentence-transformers/clip-ViT-B-32-multilingual-v1} for text) and/or BLIP-2 (\textit{Salesforce/blip2-flan-t5-xl}). We average-pool the backbone output tokens along the sequence dimension to obtain a single latent vector per sample.
Using augmentation allows joint pretraining of the attention mechanism with the GW: for the representation learning losses, we feed augmented inputs to the encoders while decoders predict non-augmented (``clean'') targets.
We train the attention mechanism only on elements of the loss where image and text are jointly provided as the input partition; for all other input partitions we use random fusion scores, as in the Simple Shapes case.
% ==================================================
\section{Additional Results}
\label{sec:supp_results}
% ==================================================
\subsection{Confusion matrix for MM-imdb classification}

\begin{figure}[t]
\centering
\includegraphics[width=0.95\linewidth]{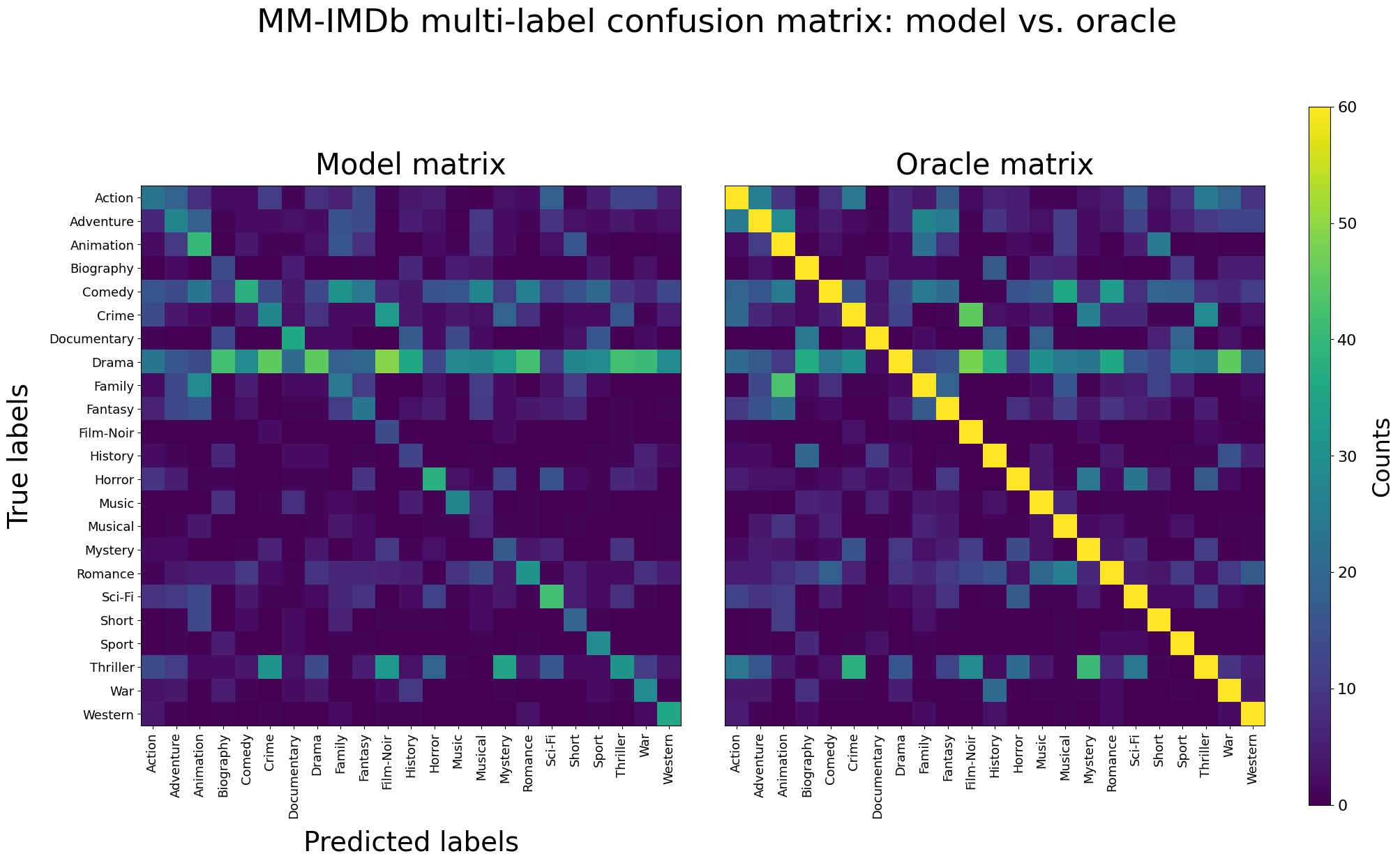}
\caption{
Multi-label confusion matrix under maximum corruption, with our MM-IMDb model trained on maximum noise. Rows indicate true target labels and columns indicate predicted labels.
}
\label{fig:confusion_matrix}
\vspace{-1em}
\end{figure}

We provide in figure \ref{fig:confusion_matrix} a confusion map of predicted vs target classes under maximum corruption, for our MM-IMDB probes with attention trained on maximum corruption. Because MM-IMDb is multi-label, off-diagonal values do not necessarily indicate mistakes, but can also be correct label predictions. For this reason, we also provide an "oracle" confusion matrix representing "perfect classification" (which is not identical to the identity matrix). Note that classes aren't equally probable: Drama, for instance, is overrepresented in the dataset. Overall, we don't notice class imbalances in the predictions, only that our model is too cautious: it generally predicts fewer labels than the true number.

\subsection{Parameter-count breakdown for robustness experiments}
\label{subsec:supp_param_counts}

We provide a full parameter breakdown both on MM-IMDb and Simple Shapes in table \ref{tab:supp_param_counts}.

\begin{table}[t]
\caption{Parameter-count breakdowns for the robustness experiments. ``Label'' denotes the parameters trained for clean-task supervision, ``Corr.'' those trained during corruption robustness learning, and ``Frozen'' the parameters kept fixed during robustness training.}
\label{tab:supp_param_counts}
\centering
\scriptsize
\setlength{\tabcolsep}{3pt}
\begin{tabular}{lrrrr}
    \toprule
    Model & Params & \shortstack{Label} & \shortstack{Corr.} & \shortstack{Frozen} \\
    \midrule
    \multicolumn{5}{l}{\textit{Simple Shapes}} \\
    GMU   & 113{,}340 & 113{,}340 & 113{,}340 & 2{,}981{,}946 \\
    DynMM & 92{,}578  & 92{,}578  & 92{,}578  & 2{,}981{,}946 \\
    Ours  & 68{,}764  & 64{,}220  & 4{,}544   & 3{,}082{,}540 \\
    \midrule
    \multicolumn{5}{l}{\textit{MM-IMDb 1.0}} \\
    DynMM & 4.48M & 4.48M & 4.48M & 196.06M \\
    GMU   & 5.01M & 5.01M & 5.01M & 196.06M \\
    Ours  & 7.98M & 4.02M & 0.06M & 204.68M \\
    \bottomrule
\end{tabular}
\end{table}

\subsection{Modality-wise generalization II: unseen modality}

As a second, stronger modality generalization task, we start once again from the frozen GW and classification probes trained on clean data. Then, we leave out one of the modalities completely during attention training. Since there are only two modalities left, we adapt the training noise schedule to have only one noisy and one clean modality, randomly determined on each sample. Then at inference, we add the left-out modality back (with 2 randomly assigned clean modality and 1 noisy one), and test whether the attention mechanism is able to recognize when this left-out modality is clean vs. noisy, and select it or tune it out (respectively).
For this experiment, we cannot include GMU and DynMM as baselines, because their gating mechanisms are designed for a fixed number of modalities. In our case, as the key layer is shared across modalities, our attentional mechanism can be flexibly extended to new modalities. To validate this transfer capability, we use random fusion scores as a no-attention baseline.

\begin{figure}[t]
  \centering
  \includegraphics[width=0.9\linewidth]{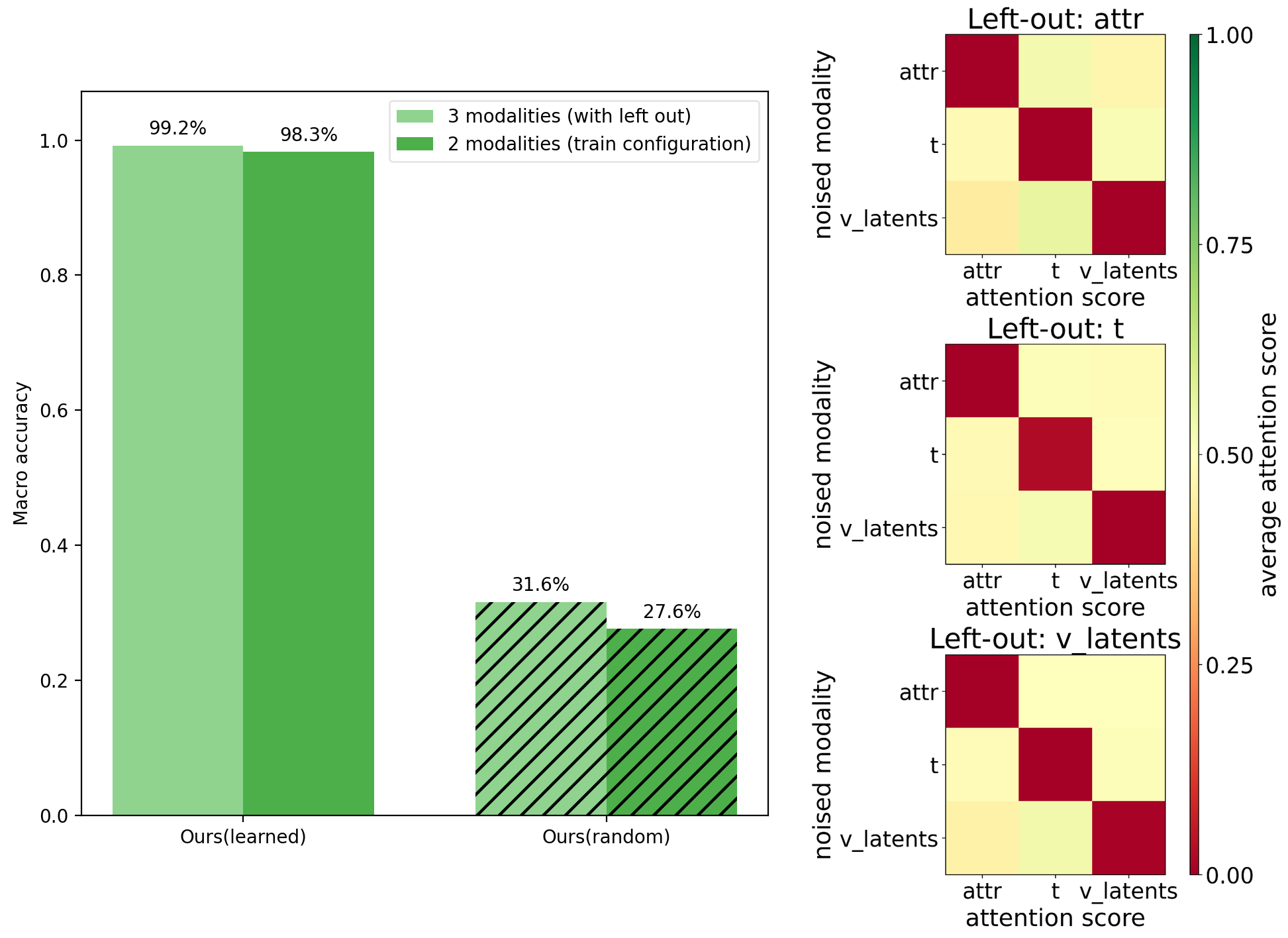}
  \caption{Modality generalization test II: unseen modality. We train the attention system on two and test on three modalities.
  \textbf{Left}: Our attention mechanism's performance vs. random fusion performance (no-attention), under both train and evaluation configurations. The bars show accuracy averaged across 3 possible left-out modalities, 5 tasks, and all combinations of noisy/clean modalities.
  \textbf{Right}: average attention scores given by our attention mechanism on the test dataset, as a function of the noised modality. We show results separately for each left-out modality, and notice that attention scores on the left-out modality are comparable to trained modalities: this means our attention mechanism perfectly generalizes to the left-out modality.}
  \label{fig:strong_att_gen}
\end{figure}

Figure~\ref{fig:strong_att_gen} shows that our attention system is indeed able to generalize to an additional modality: adding a third modality actually increases performance, because now only one of three modalities is noised (instead of one of two during training). In contrast, we see that adding a third modality only slightly increases performance for the random variant, meaning that the noise corruption remains a challenging test, even when two clean modalities are present instead of one. In addition, we analyze the attention scores given to the left-out modality, compared to the trained modalities. We see that the trained attention strategy extends to the left-out, unseen modality. This is made possible by the amodal representation learned with our broadcast loss, combined with the sharing of the key matrix across modalities.

\bibliographystyle{splncs04}
\bibliography{icann_2026}

@article{baltrusaitis2019multimodal,
  author  = {Baltru{\v{s}}aitis, Tadas and Ahuja, Chaitanya and Morency, Louis-Philippe},
  title   = {Multimodal Machine Learning: A Survey and Taxonomy},
  journal = {IEEE Transactions on Pattern Analysis and Machine Intelligence},
  volume  = {41},
  number  = {2},
  pages   = {423--443},
  year    = {2019},
  doi     = {10.1109/TPAMI.2018.2798607}
}

@inproceedings{ngiam2011multimodal,
  author    = {Ngiam, Jiquan and Khosla, Aditya and Kim, Mingyu and Nam, Juhan and Lee, Honglak and Ng, Andrew Y.},
  title     = {Multimodal Deep Learning},
  booktitle = {Proceedings of the 28th International Conference on Machine Learning},
  pages     = {689--696},
  year      = {2011}
}

@inproceedings{radford2021clip,
  author    = {Radford, Alec and Kim, Jong Wook and Hallacy, Chris and Ramesh, Aditya and Goh, Gabriel and Agarwal, Sandhini and Sastry, Girish and Askell, Amanda and Mishkin, Pamela and Clark, Jack and Krueger, Gretchen and Sutskever, Ilya},
  title     = {Learning Transferable Visual Models From Natural Language Supervision},
  booktitle = {Proceedings of the 38th International Conference on Machine Learning},
  series    = {Proceedings of Machine Learning Research},
  volume    = {139},
  pages     = {8748--8763},
  year      = {2021}
}

@article{huang2023kosmos,
  author  = {Huang, Shaohan and Dong, Li and Wang, Wenhui and Hao, Yaru and Singhal, Saksham and Ma, Shuming and Lv, Tengchao and Cui, Lei and Mohammed, Owais Khan and Patra, Barun and Liu, Qiang and Aggarwal, Kriti and Chi, Zewen and Bjorck, Johan and Chaudhary, Vishrav and Som, Subhojit and Song, Xia and Wei, Furu},
  title   = {Language Is Not All You Need: Aligning Perception with Language Models},
  journal = {arXiv preprint arXiv:2302.14045},
  year    = {2023},
  doi     = {10.48550/arXiv.2302.14045}
}

@inproceedings{bao2020gwn,
  author    = {Bao, Cong and Fountas, Zafeirios and Olugbade, Temitayo and Bianchi-Berthouze, Nadia},
  title     = {{GWN}: Multimodal Data Fusion based on the Global Workspace via Hierarchical Skip Connections and Multi-Head Attention},
  booktitle = {Proceedings of the 2020 International Conference on Multimodal Interaction},
  year      = {2020},
  doi       = {10.1145/3382507.3418849}
}

@article{goyal2021sgw,
  author  = {Goyal, Anirudh and Didolkar, Aniket and Lamb, Alex and Badola, Kartikeya and Ke, Nan Rosemary and Rahaman, Nasim and Binas, Jonathan and Blundell, Charles and Mozer, Michael C. and Bengio, Yoshua},
  title   = {Coordination Among Neural Modules Through a Shared Global Workspace},
  journal = {arXiv preprint arXiv:2103.01197},
  year    = {2021}
}

@article{dossa2024gwagent,
  author  = {Dossa, R. F. J. and Methnani, M. and others},
  title   = {Design and Evaluation of a Global Workspace Agent Embodied in a Realistic Multimodal Environment},
  journal = {Frontiers in Computational Neuroscience},
  year    = {2024}
}

@article{devillers2025gwsemi,
  author  = {Devillers, Benjamin and Maytie, Leopold and VanRullen, Rufin},
  title   = {Semi-Supervised Multimodal Representation Learning Through a Global Workspace},
  journal = {IEEE Transactions on Neural Networks and Learning Systems},
  volume  = {36},
  number  = {5},
  pages   = {7843--7857},
  year    = {2025},
  doi     = {10.1109/TNNLS.2024.3416701}
}

@inproceedings{sun2025ait,
  author    = {Sun, Yuwei},
  title     = {Associative Transformer},
  booktitle = {Proceedings of the IEEE/CVF Conference on Computer Vision and Pattern Recognition},
  year      = {2025}
}

@book{baars1988cognitive,
  author    = {Baars, Bernard J.},
  title     = {A Cognitive Theory of Consciousness},
  publisher = {Cambridge University Press},
  year      = {1988}
}

@article{dehaene1998neuronal,
  author  = {Dehaene, Stanislas and Kerszberg, Michel and Changeux, Jean-Pierre},
  title   = {A Neuronal Model of a Global Workspace in Effortful Cognitive Tasks},
  journal = {Proceedings of the National Academy of Sciences},
  volume  = {95},
  number  = {24},
  pages   = {14529--14534},
  year    = {1998},
  doi     = {10.1073/pnas.95.24.14529}
}

@incollection{dehaene2011gnw,
  author    = {Dehaene, Stanislas and Changeux, Jean-Pierre and Naccache, Lionel},
  title     = {The Global Neuronal Workspace Model of Conscious Access: From Neuronal Architectures to Clinical Applications},
  booktitle = {Characterizing Consciousness: From Cognition to the Clinic?},
  editor    = {Dehaene, Stanislas and Christen, Yves},
  pages     = {55--84},
  publisher = {Springer},
  year      = {2011},
  doi       = {10.1007/978-3-642-18015-6_4}
}

@article{vanrullen2021tins,
  author  = {VanRullen, Rufin and Kanai, Ryota},
  title   = {Deep Learning and the Global Workspace Theory},
  journal = {Trends in Neurosciences},
  volume  = {44},
  number  = {9},
  pages   = {692--704},
  year    = {2021},
  doi     = {10.1016/j.tins.2021.04.005}
}

@inproceedings{dufumier2025comm,
  author    = {Dufumier, Benoit and Castillo-Navarro, Javiera and Tuia, Devis and Thiran, Jean-Philippe},
  title     = {What to Align in Multimodal Contrastive Learning?},
  booktitle = {International Conference on Learning Representations},
  year      = {2025},
  url       = {https://openreview.net/forum?id=Pe3AxLq6Wf}
}

@article{arevalo2017gmu,
  author  = {Arevalo, John and Solorio, Thamar and Montes-y-G{\'o}mez, Manuel and Gonz{\'a}lez, Fabio A.},
  title   = {Gated Multimodal Units for Information Fusion},
  journal = {arXiv preprint arXiv:1702.01992},
  year    = {2017},
  doi     = {10.48550/arXiv.1702.01992}
}

@inproceedings{xue2023dynmm,
  author    = {Xue, Zihui and Chen, Liang-Chieh and Wei, Yunchao and Wang, Li-Jia and Chen, Xiaojie},
  title     = {Dynamic Multimodal Fusion ({DynMM})},
  booktitle = {Proceedings of the IEEE/CVF Conference on Computer Vision and Pattern Recognition Workshops},
  year      = {2023}
}

@inproceedings{Li2023IDKG,
  author    = {Li, Jiaqi and Qi, Guilin and Zhang, Chuanyi and Chen, Yongrui and Tan, Yiming and Xia, Chenlong and Tian, Ye},
  title     = {Incorporating Domain Knowledge Graph into Multimodal Movie Genre Classification with Self-Supervised Attention and Contrastive Learning},
  booktitle = {Proceedings of the 31st ACM International Conference on Multimedia},
  pages     = {3337--3345},
  year      = {2023},
  doi       = {10.1145/3581783.3612085}
}

@article{Lample2018PhraseBasedNeuralUMT,
  author  = {Lample, Guillaume and Ott, Myle and Conneau, Alexis and Denoyer, Ludovic and Ranzato, Marc'Aurelio},
  title   = {Phrase-Based \& Neural Unsupervised Machine Translation},
  journal = {arXiv preprint arXiv:1804.07755},
  year    = {2018},
  doi     = {10.48550/arXiv.1804.07755}
}

@article{Artetxe2017UnsupervisedNMT,
  author  = {Artetxe, Mikel and Labaka, Gorka and Agirre, Eneko and Cho, Kyunghyun},
  title   = {Unsupervised Neural Machine Translation},
  journal = {arXiv preprint arXiv:1710.11041},
  year    = {2017},
  doi     = {10.48550/arXiv.1710.11041}
}

@article{Maytie2024ZeroShotGW,
  author  = {Mayti{\'e}, L{\'e}opold and Devillers, Benjamin and Arnold, Alexandre and VanRullen, Rufin},
  title   = {Zero-Shot Cross-Modal Transfer of Reinforcement Learning Policies Through a Global Workspace},
  journal = {Reinforcement Learning Journal},
  volume  = {3},
  pages   = {1410--1426},
  year    = {2024}
}

@article{Maytie2025MultimodalDreaming,
  author  = {Mayti{\'e}, L{\'e}opold and Bertin Johannet, Roland and VanRullen, Rufin},
  title   = {Multimodal Dreaming: A Global Workspace Approach to World Model-Based Reinforcement Learning},
  journal = {arXiv preprint arXiv:2502.21142},
  year    = {2025}
}

@article{neverova2016moddrop,
  author  = {Neverova, Natalia and Wolf, Christian and Taylor, Graham W. and Nebout, Florian},
  title   = {{ModDrop}: Adaptive Multi-Modal Gesture Recognition},
  journal = {IEEE Transactions on Pattern Analysis and Machine Intelligence},
  volume  = {38},
  number  = {8},
  pages   = {1692--1706},
  year    = {2016},
  doi     = {10.1109/TPAMI.2015.2461544}
}

@article{wu2024mlmm_survey,
  author  = {Wu, Renjie and Wang, Hu and Chen, Hsiang-Ting and Carneiro, Gustavo},
  title   = {Deep Multimodal Learning with Missing Modality: A Survey},
  journal = {arXiv preprint arXiv:2409.07825},
  year    = {2024},
  doi     = {10.48550/arXiv.2409.07825}
}

@inproceedings{ma2021smil,
  author    = {Ma, Mengmeng and Ren, Jian and Zhao, Long and Tulyakov, Sergey and Wu, Cathy and Peng, Xi},
  title     = {{SMIL}: Multimodal Learning with Severely Missing Modality},
  booktitle = {Proceedings of the AAAI Conference on Artificial Intelligence},
  volume    = {35},
  number    = {3},
  pages     = {2302--2310},
  year      = {2021}
}

@inproceedings{strubell2019energy,
  author    = {Strubell, Emma and Ganesh, Ananya and McCallum, Andrew},
  title     = {Energy and Policy Considerations for Deep Learning in {NLP}},
  booktitle = {Proceedings of the 57th Annual Meeting of the Association for Computational Linguistics},
  pages     = {3645--3650},
  year      = {2019},
  doi       = {10.18653/v1/P19-1355}
}

@article{bommasani2021foundation,
  author  = {Bommasani, Rishi and Hudson, Drew A. and Adeli, Ehsan and Altman, Russ and Arora, Simran and von Arx, Sydney and Bernstein, Michael S. and Bohg, Jeannette and Bosselut, Antoine and Brunskill, Emma and others},
  title   = {On the Opportunities and Risks of Foundation Models},
  journal = {arXiv preprint arXiv:2108.07258},
  year    = {2021},
  doi     = {10.48550/arXiv.2108.07258}
}

@article{qiu2024benchmarking,
  author  = {Qiu, Jielin and Zhu, Yi and Shi, Xingjian and Wenzel, Florian and Tang, Zhiqiang and Zhao, Ding and Li, Bo and Li, Mu},
  title   = {Benchmarking Robustness of Multimodal Image-Text Models under Distribution Shift},
  journal = {arXiv preprint arXiv:2212.08044},
  year    = {2024},
  doi     = {10.48550/arXiv.2212.08044}
}
% ==================================================

% Restore normal label behavior.
\makeatletter
\let\label\origlabel
\makeatother

% ==================================================
% Single bibliography for main + supplement
% ==================================================

\clearpage
\realbibliographystyle{splncs04}
\realbibliography{icann_2026}

\end{document}